# Learning Algorithms for Keyphrase Extraction

*Peter D. Turney*
*Institute for Information Technology*
*National Research Council of Canada*
*Ottawa, Ontario, Canada, K1A 0R6*
*peter.turney@iit.nrc.ca*
*Phone: 613-993-8564*
*Fax: 613-952-7151*

**Abstract**

Many academic journals ask their authors to provide a list of about five to fifteen *keywords*, to appear on the first page of each article. Since these key words are often phrases of two or more words, we prefer to call them *keyphrases*. There is a wide variety of tasks for which keyphrases are useful, as we discuss in this paper. We approach the problem of automatically extracting keyphrases from text as a supervised learning task. We treat a document as a set of phrases, which the learning algorithm must learn to classify as positive or negative examples of keyphrases. Our first set of experiments applies the C4.5 decision tree induction algorithm to this learning task. We evaluate the performance of nine different configurations of C4.5. The second set of experiments applies the *GenEx* algorithm to the task. We developed the GenEx algorithm specifically for automatically extracting keyphrases from text. The experimental results support the claim that a custom-designed algorithm (GenEx), incorporating specialized procedural domain knowledge, can generate better keyphrases than a general-purpose algorithm (C4.5). Subjective human evaluation of the keyphrases generated by Extractor suggests that about 80% of the keyphrases are acceptable to human readers. This level of performance should be satisfactory for a wide variety of applications.

**Keyphrases:** machine learning, summarization, indexing, keywords, keyphrase extraction.

# Learning Algorithms for Keyphrase Extraction

## 1. Introduction

Many journals ask their authors to provide a list of *keywords* for their articles. We call these *keyphrases*, rather than keywords, because they are often phrases of two or more words, rather than single words. We define a *keyphrase list* as a short list of phrases (typically five to fifteen noun phrases) that capture the main topics discussed in a given document. This paper is concerned with the automatic extraction of keyphrases from text.

Keyphrases are meant to serve multiple goals. For example, (1) when they are printed on the first page of a journal article, the goal is summarization. They enable the reader to quickly determine whether the given article is in the reader's fields of interest. (2) When they are printed in the cumulative index for a journal, the goal is indexing. They enable the reader to quickly find a relevant article when the reader has a specific need. (3) When a search engine form has a field labelled *keywords*, the goal is to enable the reader to make the search more precise. A search for documents that match a given query term in the *keyword* field will yield a smaller, higher quality list of hits than a search for the same term in the full text of the documents. Keyphrases can serve these diverse goals and others, because the goals share the requirement for a short list of phrases that captures the main topics of the documents.

We define *automatic keyphrase extraction* as the automatic selection of important, topical phrases from within the body of a document. Automatic keyphrase extraction is a special case of the more general task of *automatic keyphrase generation*, in which the generated phrases do not necessarily appear in the body of the given document. Section 2 discusses criteria for measuring the performance of automatic keyphrase extraction algorithms. In the experiments in this paper, we measure the performance by comparing machine-generated keyphrases with human-generated keyphrases. In our document collections, an average of about 75% of the author's keyphrases appear somewhere in the body of the corresponding document. Thus, an ideal keyphrase extraction algorithm could (in principle) generate phrases that match up to 75% of the author's keyphrases.

There is a need for tools that can automatically create keyphrases. Although keyphrases are very useful, only a small minority of the many documents that are available on-line today have keyphrases. There are already some commercial software products that use automatic keyphrase extraction algorithms. For example, Microsoft uses automatic keyphrase extraction in Word 97, to fill the *Keywords* field in the document metadata template (metadata is meta-information for document management).[1] Verity uses automatic keyphrase extraction in Search 97, their search engine product line. In Search 97, keyphrases are highlighted in bold to facilitate skimming through a list of search results.[2] Tetranet uses automatic keyphrase extraction in their Metabot product, which is designed for maintaining metadata for web pages. Tetranet also uses automatic keyphrase extraction in their Wisebot product, which builds an index for a web site.[3]

Although the applications for keyphrases mentioned above share the requirement for a short list of phrases that captures the main topics of the documents, the precise size of the list will vary, depending on the particular application and the inclinations of the users. Therefore the algorithms that we discuss allow the users to specify the desired number of phrases.

We discuss related work by other researchers in Section 3. The most closely related work involves the problem of *automatic index generation* (Fagan, 1987; Salton, 1988; Ginsberg, 1993; Nakagawa, 1997; Leung and Kan, 1997). One difference between keyphrase extraction

1. To access the metadata template in Word 97, select *File* and then *Properties*. To automatically fill the *Keywords* field, select *Tools* and then *AutoSummarize*. (This is not obvious from the Word 97 documentation.) Microsoft and Word 97 are trademarks or registered trademarks of Microsoft Corporation.

2. Microsoft and Verity use proprietary techniques for keyphrase extraction. It appears that their techniques do not involve machine learning. Verity and Search 97 are trademarks or registered trademarks of Verity Inc.

3. Tetranet has licensed our keyphrase extraction software for use in their products. Tetranet, Metabot, and Wisebot are trademarks or registered trademarks of Tetranet Software. For experimental comparisons of Word 97 and Search 97 with our own work, see Turney (1997, 1999).

and index generation is that, although keyphrases may be used in an index, keyphrases have other applications, beyond indexing. Another difference between a keyphrase list and an index is length. Because a keyphrase list is relatively short, it must contain only the most important, topical phrases for a given document. Because an index is relatively long, it can contain many less important, less topical phrases. Also, a keyphrase list can be read and judged in seconds, but an index might never be read in its entirety. Automatic keyphrase extraction is thus a more demanding task than automatic index generation.

Keyphrase extraction is also distinct from *information extraction*, the task that has been studied in depth in the *Message Understanding Conferences* (MUC-3, 1991; MUC-4, 1992; MUC-5, 1993; MUC-6, 1995). Information extraction involves extracting specific types of task-dependent information. For example, given a collection of news reports on terrorist attacks, information extraction involves finding specific kinds of information, such as the name of the terrorist organization, the names of the victims, and the type of incident (e.g., kidnapping, murder, bombing). In contrast, keyphrase extraction is not specific. The goal in keyphrase extraction is to produce topical phrases, for any type of factual, prosaic document.

We approach automatic keyphrase extraction as a supervised learning task. We treat a document as a set of phrases, which must be classified as either positive or negative examples of keyphrases. This is the classical machine learning problem of *learning from examples*. In Section 5, we describe how we apply the C4.5 decision tree induction algorithm to this task (Quinlan, 1993). There are several unusual aspects to this classification problem. For example, the positive examples constitute only 0.2% to 2.4% of the total number of examples. C4.5 is typically applied to more balanced class distributions.

The experiments in this paper use five collections of documents, with a combined total of 652 documents. The collections are presented in Section 4. In our first set of experiments (Section 6), we evaluate nine different ways to apply C4.5. In preliminary experiments with the training documents, we found that *bagging* seemed to improve the performance of C4.5 (Breiman, 1996a, 1996b; Quinlan, 1996). Bagging works by generating many different decision trees and allowing them to vote on the classification of each example. We experimented with different numbers of trees and different techniques for sampling the training data. The experiments support the hypothesis that bagging improves the performance of C4.5 when applied to automatic keyphrase extraction.

During our experiments with C4.5, we came to believe that a specialized algorithm, developed specifically for learning to extract keyphrases, might achieve better results than a general-purpose learning algorithm, such as C4.5. Section 7 introduces the *GenEx* algorithm. *GenEx* is a hybrid of the *Genitor* steady-state genetic algorithm (Whitley, 1989) and the *Extractor* parameterized keyphrase extraction algorithm (Turney, 1997, 1999).[4] Extractor works by assigning a numerical score to the phrases in the input document. The final output of Extractor is essentially a list of the highest scoring phrases. The behaviour of the scoring function is determined by a dozen numerical parameters. Genitor tunes the setting of these parameters, to maximize the performance of Extractor on a given set of training examples.

The second set of experiments (Section 8) supports the hypothesis that a specialized algorithm (GenEx) can generate better keyphrases than a general-purpose algorithm (C4.5). Both algorithms incorporate significant amounts of domain knowledge, but we avoided embedding specialized procedural knowledge in our application of C4.5. It appears that some degree of specialized procedural knowledge is necessary for automatic keyphrase extraction.

The third experiment (Section 9) looks at subjective human evaluation of the quality of the keyphrases produced by GenEx. On average, about 80% of the automatically generated keyphrases are judged to be acceptable and about 60% are judged to be good.

Section 10 discusses the experimental results and Section 11 presents our plans for future work. We conclude (in Section 12) that GenEx is performing at a level that is suitable for

4. Extractor is an Official Mark of the National Research Council of Canada. Patent applications have been submitted for Extractor.

many practical applications.

## 2. Measuring the Performance of Keyphrase Extraction Algorithms

We measure the performance of keyphrase extraction algorithms by the number of matches between the machine-generated phrases and the human-generated phrases. A handmade keyphrase *matches* a machine-generated keyphrase when they correspond to the same sequence of stems. A stem is what remains when we remove the suffix from a word. By this definition, “neural networks” matches “neural network”, but it does not match “networks”. The order in the sequence is important, so “helicopter skiing” does not match “skiing helicopter”.

The Porter (1980) and Lovins (1968) stemming algorithms are the two most popular algorithms for stemming English words. Both algorithms use heuristic rules to remove or transform English suffixes. The Lovins stemmer is more aggressive than the Porter stemmer. That is, the Lovins stemmer is more likely to recognize that two words share the same stem, but it is also more likely to incorrectly map two distinct words to the same stem (Krovetz, 1993). We have found that aggressive stemming is better for keyphrase extraction than conservative stemming. In our experiments, we have used an aggressive stemming algorithm that we call the Iterated Lovins stemmer. The algorithm repeatedly applies the Lovins stemmer, until the word stops changing. Iterating in this manner will necessarily increase (or leave unchanged) the aggressiveness of any stemmer. Table 1 shows some examples of the behaviour of the three stemming algorithms.[5]

We may view keyphrase extraction as a classification problem. The task is to classify

5. We used an implementation of the Porter (1980) stemming algorithm written in Perl, by Jim Richardson, at the University of Sydney, Australia. This implementation includes some extensions to Porter’s original algorithm, to handle British spelling. It is available at http://www.maths.usyd.edu.au:8000/jimr.html. For the Lovins (1968) stemming algorithm, we used an implementation written in C, by Linh Huynh. This implementation is part of the MG (Managing Gigabytes) search engine, which was developed by a group of people in Australia and New Zealand. The MG code is available at http://www.cs.mu.oz.au/mg/.

Table 1: Samples of the behaviour of three different stemming algorithms.

| Word | Porter Stem | Lovins Stem | Iterated Lovins Stem |
|---|---|---|---|
| believes | believ | belief | belief |
| belief | belief | belief | belief |
| believable | believ | belief | belief |
| jealousness | jealous | jeal | jeal |
| jealousy | jealousi | jealous | jeal |
| police | polic | polic | pol |
| policy | polici | polic | pol |
| assemblies | assembli | assembl | assembl |
| assembly | assembli | assemb | assemb |
| probable | probabl | prob | prob |
| probability | probabl | prob | prob |
| probabilities | probabl | probabil | probabil |

each word or phrase in the document into one of two categories: either it is a keyphrase or it is not a keyphrase. We evaluate automatic keyphrase extraction by the degree to which its classifications correspond to human-generated classifications. Our performance measure is *precision* (the number of matches divided by the number of machine-generated keyphrases), using a variety of cut-offs for the number of machine-generated keyphrases.

## 3. Related Work

Although there are several papers that discuss automatically extracting important phrases, as far as we know, we are the first to treat this problem as supervised learning from examples. Krulwich and Burkey (1996) use heuristics to extract keyphrases from a document. The heuristics are based on syntactic clues, such as the use of italics, the presence of phrases in section headers, and the use of acronyms. Their motivation is to produce phrases for use as features when automatically classifying documents. Their algorithm tends to produce a relatively large list of phrases, with low precision. Muñoz (1996) uses an unsupervised learning algorithm to discover two-word keyphrases. The algorithm is based on Adaptive Resonance Theory (ART) neural networks. Muñoz’s algorithm tends to produce a large list of phrases, with low precision. Also, the algorithm is not applicable to one-word or more-than-two-word keyphrases. Steier and Belew (1993) use the mutual information statistic to discover two-word keyphrases. This approach has the same limitations as Muñoz (1996), when considered as a keyphrase extraction algorithm: it produces a low precision list of two-word phrases.

In the time since this paper was submitted for publication, Frank *et al.* (1999) have implemented a system, *Kea*, which builds on our work (Turney, 1997, 1999). It treats keyphrase extraction as a supervised learning problem, but it uses a Bayesian approach instead of a genetic algorithm approach. Their experiments indicate that Kea and GenEx have statistically equivalent levels of performance. The same group (Gutwin *et al.*, 1999) has evaluated Kea as a component in a new kind of search engine, *Keyphind*, designed specially to support browsing. Their experiments suggest that certain kinds of tasks are much easier with Keyphind than with conventional search engines. The Keyphind interface is somewhat similar to the interface of Tetranet's Wisebot.

Several papers explore the task of producing a summary of a document by extracting key sentences from the document (Luhn, 1958; Edmundson, 1969; Marsh *et al.*, 1984; Paice, 1990; Paice and Jones, 1993; Johnson *et al.*, 1993; Salton *et al.*, 1994; Kupiec *et al.*, 1995; Brandow *et al.*, 1995; Jang and Myaeng, 1997). This task is similar to the task of keyphrase extraction, but it is more difficult. The extracted sentences often lack cohesion because anaphoric references are not resolved (Johnson *et al.*, 1993; Brandow *et al.*, 1995). *Anaphors* are pronouns (e.g., “it”, “they”), definite noun phrases (e.g., “the car”), and demonstratives (e.g., “this”, “these”) that refer to previously discussed concepts. When a sentence is extracted out of the context of its neighbouring sentences, it may be impossible or very difficult for the reader of the summary to determine the referents of the anaphors. Johnson *et al.* (1993) attempt to automatically resolve anaphors, but their system tends to produce overly long summaries. Keyphrase extraction avoids this problem because anaphors (by their nature) are not keyphrases. Also, a list of keyphrases has no structure; unlike a list of sentences, a list of keyphrases can be randomly permuted without significant consequences.

Most of these papers on summarization by sentence extraction describe algorithms that are based on manually derived heuristics. The heuristics tend to be effective for the intended domain, but they often do not generalize well to a new domain. Extending the heuristics to a new domain involves a significant amount of manual work. A few of the papers describe learning algorithms, which can be trained by supplying documents with associated target summaries (Kupiec *et al.*, 1995; Jang and Myaeng, 1997). Learning algorithms can be extended to new domains with less work than algorithms that use manually derived heuristics. However, there is still some manual work involved, because the training summaries must be composed of sentences that appear in the document, which means that standard author-supplied abstracts are not suitable. An advantage of keyphrase extraction is that standard author-supplied keyphrases are suitable for training a learning algorithm, because the majority of such keyphrases appear in the bodies of the corresponding documents.

Another body of related work addresses the task of *information extraction*. An information extraction system seeks specific information in a document, according to predefined template. The template is specific to a given topic area. For example, if the topic area is news reports of terrorist attacks, the template might specify that the information extraction system should identify (i) the terrorist organization involved in the attack, (ii) the victims of the attack, (iii) the type of attack (kidnapping, murder, etc.), and other information of this type. ARPA has sponsored a series of *Message Understanding Conferences* (MUC-3, 1991; MUC-4, 1992; MUC-5, 1993; MUC-6, 1995), where information extraction systems are evaluated with corpora in various topic areas, including terrorist attacks and corporate mergers.

Most information extraction systems are manually built for a single topic area, which requires a large amount of expert labour. The highest performance at the Fifth Message Understanding Conference (MUC-5, 1993) was achieved at the cost of two years of intense programming effort. However, recent work has demonstrated that a learning algorithm can perform as well as a manually constructed system (Soderland and Lehnert, 1994). Soderland and Lehnert (1994) use decision tree induction as the learning component in their informa-tion extraction system. In Soderland and Lehnert’s (1994) system, each slot in a template is handled by a group of decision trees that have been trained specially for that slot. The decision trees are based on syntactical features of the text, such as the presence of certain words.

Information extraction and keyphrase extraction are at opposite ends of a continuum that ranges from detailed, specific, and domain-dependent (information extraction) to condensed, general, and domain-independent (keyphrase extraction). The different ends of this continuum require substantially different algorithms. However, there are intermediate points on this continuum. An example is the task of identifying corporate names in business news. This task was introduced in the Sixth Message Understanding Conference (MUC-6, 1995), where it was called the *Named Entity Recognition* task. The best system used hand-crafted linguistic rules to recognize named entities (Krupka, 1995).

Other related work addresses the problem of automatically creating an index (Sparck Jones, 1973; Field, 1975; Fagan, 1987; Salton, 1988; Croft *et al.*, 1991; Ginsberg, 1993; Nakagawa, 1997; Leung and Kan, 1997). Leung and Kan (1997) provide a good survey of this work. There are two general classes of indexes: indexes that are intended for human readers to browse (often called *back-of-book* indexes) and indexes that are intended for use with information retrieval software (*search engine* indexes). Search engine indexes are not suitable for human browsing, since they usually index every occurrence of every word (excluding stop words) in the document collection. Back-of-book indexes tend to be much smaller, since they index only important occurrences of interesting words and phrases. The older work on automatically creating an index (Sparck Jones, 1973; Field, 1975; Fagan, 1987) is concerned with building search engine indexes, not with building back-of-book indexes (Salton, 1988; Nakagawa, 1997; Leung and Kan, 1997).

Since we are interested in keyphrases for human browsing, back-of-book indexes are more relevant than search engine indexes. Leung and Kan (1997) address the problem of learning to assign index terms from a controlled vocabulary. This involves building a statisti-cal model for each index term in the controlled vocabulary. The statistical model attempts to capture the syntactic properties that distinguish documents for which the given index term is appropriate from documents for which it is inappropriate. Their results are interesting, but the use of a controlled vocabulary makes it difficult to compare their work with the algorithms we examine here. It is also worth noting that a list of controlled index terms must grow every year, as the body of literature grows, so Leung and Kan's (1997) software would need to be continuously trained.

Nakagawa (1997) automatically extracts simple and compound nouns from technical manuals, to create back-of-book indexes. Each compound noun is scored using a formula that is based on the frequency of its component nouns in the given document. In his experiments, Nakagawa (1997) evaluates his algorithm by comparing human-generated indexes to machine-generated indexes. However, Nakagawa's (1997) human-generated indexes were generated with the assistance of his algorithm, which tends to bias the results.

One feature that distinguishes a back-of-book index from a keyphrase list is length. As Nakagawa (1997) observes, a document is typically assigned $10^0 \sim 10^1$ keyphrases, but a back-of-book index typically contains $10^2 \sim 10^3$ index terms. Also, keyphrases are usually intended to cover the whole document, but index terms are intended to cover only a small part of a document. Another distinguishing feature is that a sophisticated back-of-book index is not simply an alphabetical list of terms. There is often a hierarchical structure, where a major index term is followed by an indented list of related minor index terms.

## 4. The Corpora

The experiments in this paper are based on five different document collections. For each document, there is a target set of keyphrases, generated by hand. Some basic statistics for the five corpora are presented in Table 2.

For the Journal Articles corpus, we selected 75 journal articles from five different journals. Three of the journals are about cognition (*Psycoloquy, The Neuroscientist, Behavioral*

Table 2: Some statistics for each of the five collections.

| Corpus Name | Number of Documents | Average Number of … ± Standard Deviation | | | Percentage of Keyphrases in Full Text |
|---|---|---|---|---|---|
| | | Keyphrases per Document | Words per Keyphrase | Words per Document | |
| Journal Articles | 75 | 7.5 ± 2.8 | 1.6 ± 0.7 | 10,781 ± 7,807 | 81.6% |
| Email Messages | 311 | 4.9 ± 4.3 | 1.8 ± 1.0 | 376 ± 561 | 97.9% |
| Aliweb Web Pages | 90 | 6.0 ± 3.0 | 1.2 ± 0.5 | 949 ± 2603 | 69.0% |
| NASA Web Pages | 141 | 4.7 ± 2.0 | 1.9 ± 0.9 | 466 ± 102 | 65.3% |
| FIPS Web Pages | 35 | 9.0 ± 3.5 | 2.0 ± 1.1 | 7025 ± 6385 | 78.2% |

*& Brain Sciences Preprint Archive*), one is about the hotel industry (*Journal of the International Academy of Hospitality Research*), and one is about chemistry (*Journal of Computer-Aided Molecular Design*). The full text of every article is available on the web. The authors supplied keyphrases for their articles. For the Email Messages corpus, we collected 311 email messages from six NRC employees. A university student created the keyphrases for the messages. For the Aliweb corpus, we collected 90 web pages using the Aliweb search engine, a public search engine provided by NEXOR.[6] Aliweb has web-based fill-in form, with a field for keyphrases, where people may enter URLs to add to Aliweb. The keyphrases are stored in the Aliweb index, along with the URLs. For the NASA corpus, we collected 141 web pages from NASA's Langley Research Center.[7] Each page includes a list of keyphrases. For the FIPS corpus, we gathered 35 web pages from the US government's Federal Information Processing Standards (FIPS).[8] Each document includes a list of keyphrases.

We would like our learning algorithms to be able to perform well even when the testing data are significantly different from the training data. In a real-world application, it would be inconvenient if the learning algorithm required re-training for each new type of document. Therefore our experiments do not use a random split of the documents into training and testing sets. Instead, we designed the experiments to test the ability of the learning algorithms to

6. The URL is http://www.nexor.com/public/aliweb/search/doc/form.html.

7. Available at http://tag-www.larc.nasa.gov/tops/tops_text.html.

8. Available at http://www.itl.nist.gov/div897/pubs/.

generalize to new data.

In our preliminary experiments, we found that the main factor influencing generalization was the average length of the documents in the training set, compared to the testing set. In a real-world application, it would be reasonable to have two different learned models, one for short documents and one for long documents. As Table 3 shows, we selected part of the Journal Article corpus to train the learning algorithms to handle long documents and part of the Email Message corpus to train the learning algorithms to handle short documents. During testing, we used the training corpus that was most similar to the given testing corpus, with respect to document lengths.

Table 3: The correspondence between testing and training data.

| Testing Corpus | | ↔ | Corresponding Training Corpus | |
|---|---|---|---|---|
| Name | Number of Documents | | Name | Number of Documents |
| Journal Articles (Testing Subset) | 20 | ↔ | Journal Article (Training Subset) | 55 |
| Email Messages (Testing Subset) | 76 | ↔ | Email Messages (Training Subset) | 235 |
| Aliweb Web Pages | 90 | ↔ | Email Messages (Training Subset) | 235 |
| NASA Web Pages | 141 | ↔ | Email Messages (Training Subset) | 235 |
| FIPS Web Pages | 35 | ↔ | Journal Article (Training Subset) | 55 |

The method for applying C4.5 to keyphrase extraction (Section 5) and the GenEx algorithm (Section 7) were developed using only the training subsets of the Journal Article and Email Message corpora (Table 3). The other three corpora were only acquired *after* the completion of the design of the method for applying C4.5 and the design of the GenEx algorithm. This practice ensures that there is no risk that C4.5 and GenEx have been tuned to the testing data.

## 5. Applying C4.5 to Keyphrase Extraction

In the first set of experiments, we used the C4.5 decision tree induction algorithm (Quinlan, 1993) to classify phrases as positive or negative examples of keyphrases. In this section, we describe the feature vectors, the settings we used for C4.5's parameters, the bagging proce-dure, and the method for sampling the training data.

The task of supervised learning is to learn how to assign *cases* (or *examples*) to *classes*. For keyphrase extraction, a case is a candidate phrase, which we wish to classify as a positive or negative example of a keyphrase. We classify a case by examining its *features*. A feature can be any property of a case that is relevant for determining the class of the case. C4.5 can handle real-valued features, integer-valued features, and features with values that range over an arbitrary, fixed set of symbols. C4.5 takes as input a set of *training* data, in which cases are represented as feature vectors. In the training data, a *teacher* must assign a class to each feature vector (hence *supervised* learning). C4.5 generates as output a decision tree that models the relationships among the features and the classes (Quinlan, 1993).

A *decision tree* is a rooted tree in which the internal vertices are labelled with tests on feature values and the leaf vertices are labelled with classes. The edges that leave an internal vertex are labelled with the possible outcomes of the test associated with that vertex. For example, a feature might be, “the number of words in the given phrase,” and a test on a feature value might be, “the number of words in the given phrase is less than two,” which can have the outcomes “true” or “false”. A case is classified by beginning at the root of the tree and following a path to a leaf in the tree, based on the values of the features of the case. The label on the leaf is the predicted class for the given case.

We converted the documents into sets of feature vectors by first making a list of all phrases of one, two, or three consecutive non-stop words that appear in a given document, with no intervening punctuation. We used the Iterated Lovins stemmer to find the stemmed form of each of these phrases. For each unique stemmed phrase, we generated a feature vector, as described in Table 4.

C4.5 has access to nine features (features 3 to 11) when building a decision tree. The leaves of the tree predict `class` (feature 12). When a decision tree predicts that the `class` of a vector is 1, then the phrase `whole_phrase` is a keyphrase, according to the tree. This phrase is suitable for output for a human reader. We used the stemmed form of the phrase, stemmed_phrase, for evaluating the performance of the tree. In our preliminary experiments, we evaluated 110 different features, before we settled on the features in Table 4.

Table 4: A description of the feature vectors used by C4.5.

| | Name of Feature | Description of Feature | C4.5 Type |
|---|---|---|---|
| 1 | stemmed_phrase | the stemmed form of a phrase, for matching with human-generated phrases | ignore |
| 2 | whole_phrase | the most frequent whole (unstemmed) phrase corresponding to stemmed_phrase, for output and for calculating features 8 to 11 | ignore |
| 3 | num_words_phrase | the number of words in stemmed_phrase, ranging from one to three | continuous |
| 4 | first_occur_phrase | the first occurrence of stemmed_phrase, normalized by dividing by the number of words in the document (including stop words) | continuous |
| 5 | first_occur_word | the first occurrence of the earliest occurring single stemmed word in stemmed_phrase, normalized by dividing by the number of words in the document (including stop words) | continuous |
| 6 | freq_phrase | the frequency of stemmed_phrase, normalized by dividing by the number of words in the document (including stop words) | continuous |
| 7 | freq_word | the frequency of the most frequent single stemmed word in stemmed_phrase, normalized by dividing by the number of words in the document (including stop words) | continuous |
| 8 | relative_length | the relative length of whole_phrase, calculated as the number of characters in whole_phrase, divided by the average number of characters in all candidate phrases | continuous |
| 9 | proper_noun | is whole_phrase a proper noun, based on the capitalization of whole_phrase? | 0, 1 |
| 10 | final_adjective | does whole_phrase end in a final adjective, based on the suffix of whole_phrase? | 0, 1 |
| 11 | common_verb | does whole_phrase contain a common verb, based on a list of common verbs? | 0, 1 |
| 12 | class | is stemmed_phrase a keyphrase, based on match with stemmed form of human-generated keyphrases? | 0, 1 |

Table 5 shows the number of feature vectors that were generated for each corpus. In total, we had more than 192,000 vectors for training and more than 168,000 vectors for testing. The large majority of these vectors were negative examples of keyphrases (class 0).

In a real-world application, the user would want to specify the desired number of output keyphrases for a given document. However, a standard decision tree does not let the user con-trol the number of feature vectors that are classified as belonging in class 1. Therefore we ran C4.5 with the `-p` option, which generates *soft-threshold* decision trees (Carter and Catlett, 1987; Quinlan, 1987, 1990, 1993). Soft-threshold decision trees can generate a probability estimate for the class of each vector. For a given document, if the user specifies that *K* key-phrases are desired, then we select the *K* vectors that have the highest estimated probability of being in class 1.

Table 5: The number of feature vectors for each corpus.

| Train/Test | Corpus Name | Number of Documents | Total Number of Vectors | Average Vectors Per Document | Percent Class 1 |
|---|---|---|---|---|---|
| Training | Journal | 55 | 158,240 | 2,877 | 0.20% |
| | Email | 235 | 34,378 | 146 | 2.44% |
| | All | 290 | 192,618 | 664 | 0.60% |
| Testing | Journal | 20 | 23,751 | 1,188 | 0.53% |
| | Email | 76 | 11,065 | 146 | 2.40% |
| | Aliweb | 90 | 26,752 | 297 | 1.08% |
| | NASA | 141 | 38,920 | 276 | 1.15% |
| | FIPS | 35 | 67,777 | 1,936 | 0.33% |
| | All | 362 | 168,265 | 465 | 0.80% |

In addition to the `-p` option, we also used `-c100` and `-m1` (Quinlan, 1993). These two options maximize the bushiness of the trees. In our preliminary experiments, we found that these parameter settings appear to work well when used in conjunction with bagging. *Bagging* involves generating many different decision trees and allowing them to vote on the classification of each example (Breiman, 1996a, 1996b; Quinlan, 1996). In general, decision tree induction algorithms have low bias but high variance. Bagging multiple trees tends to improve performance by reducing variance. Bagging appears to have relatively little impact on bias.

Because we used soft-threshold decision trees, we combined their probability estimates by averaging them, instead of voting. In preliminary experiments with the training documents, we obtained good results by bagging 50 decision trees. Adding more trees had no significant effect.

The standard approach to bagging is to randomly sample the training data, using sampling with replacement (Breiman, 1996a, 1996b; Quinlan, 1996). In preliminary experiments with the training data, we achieved good performance by training each of the 50 decision trees with a random sample of 1% of the training data.

The standard approach to bagging is to ignore the class when sampling, so the distribution of classes in the sample tends to correspond to the distribution in the training data as a whole. In Table 5, we see that the positive examples constitute only 0.2% to 2.4% of the total number of examples. To compensate for this, we modified the random sampling procedure so that 50% of the sampled examples were in class 0 and the other 50% were in class 1. This appeared to improve performance in preliminary experiments on the training data. This strategy is called *stratified sampling* (Deming, 1978; Buntine, 1989; Catlett, 1991; Kubat *et al.*, 1998). Kubat *et al.* (1998) found that stratified sampling significantly improved the performance of C4.5 on highly skewed data, but Catlett (1991) reported mixed results.

Boosting is another popular technique for combining multiple decision trees (Freund and Schapire, 1996; Quinlan, 1996; Maclin and Opitz, 1997). We chose to use bagging instead of boosting, because the modifications to bagging that we use here (averaging soft-threshold decision trees and stratified sampling) are simpler to apply to the bagging algorithm than to the more complicated boosting algorithm. We believe that analogous modifications would be required for boosting to perform well on this task.

## 6. Experiment 1: Learning to Extract Keyphrases with C4.5

This section presents four experiments with C4.5. In Experiment 1A, we establish a baseline for the performance of C4.5, using the configuration described in Section 5. We bag 50 trees, generated by randomly sampling 1% of the training data, with equal numbers of samples from the two classes (keyphrase and non-keyphrase). In Experiment 1B, we vary the number of trees. In Experiment 1C, we vary the ratio of the classes. In Experiment 1D, we vary the size of the random samples.

In the baseline configuration of C4.5, we bag 50 trees, where each tree is trained on a random sample of 1% of the training data, with equal samples from both classes. The performance is measured by the average precision for each corpus, when the desired number of phrases is set to 5, 7, 9, 11, 13, and 15. The average precision for a corpus is the sum of the average precision for each document, divided by the number of documents. The precision for an individual document is the number of matches between the human-generated keyphrases and the machine-generated keyphrases, divided by the desired number of phrases. Matches are determined using the Iterated Lovins stemmer.

Figure 1 shows the baseline performance of C4.5 at various cut-offs for the desired number of extracted keyphrases. The plots show the precision for the testing data only. It appears that the e-mail model generalizes relatively well to the Aliweb and NASA corpora, but the journal model does not generalize well to the FIPS corpus (see Table 3).[9]

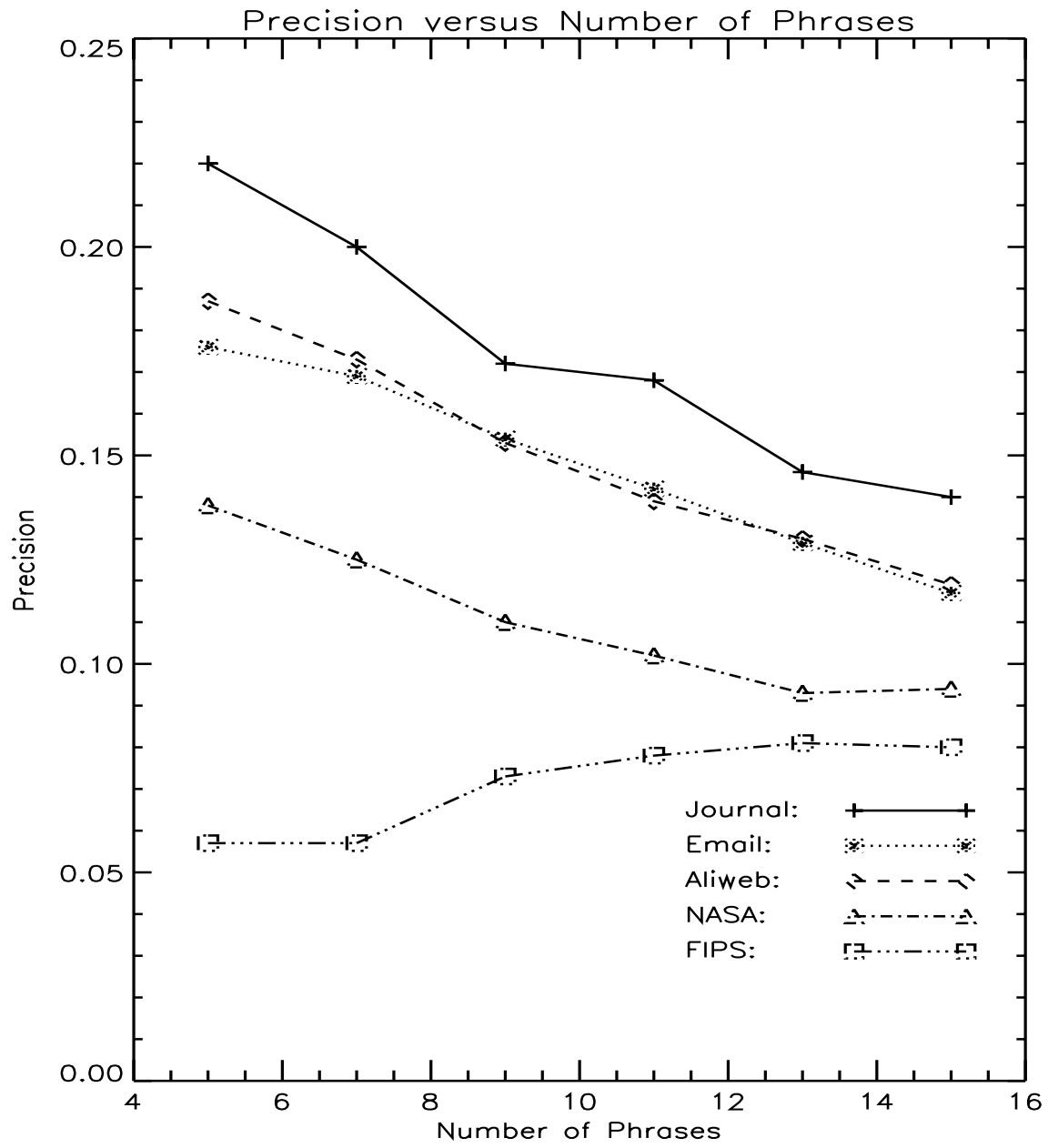

Figure 1: Experiment 1A: The baseline precision of C4.5.

9. The increasing slope of the curve for the FIPS corpus appears to be a statistical anomaly. Figure 5 shows a flatter slope for FIPS in Experiment 1C, in which we modify the stratified sampling procedure, although the slope is still somewhat unusual.

Table 6 shows the time required to train and test the baseline configuration of C4.5. The time includes generating the feature vectors and randomly sampling the vectors.[10]

Table 6: Experiment 1A: Training and testing time for the baseline configuration of C4.5.

| Train/Test | Corpus Name | Number of Documents | Total Time in Seconds * | Average Time in Seconds ** |
|---|---|---|---|---|
| Training | Journal | 55 | 250 | 60 |
| | Email | 235 | 89 | 33 |
| | | | Total Time in Seconds † | Average Time in Seconds ‡ |
| Testing | Journal | 20 | 37 | 1.9 |
| | Email | 76 | 106 | 1.4 |
| | Aliweb | 90 | 130 | 1.4 |
| | NASA | 141 | 200 | 1.4 |
| | FIPS | 35 | 86 | 2.5 |

* Total time for one corpus and all fifty trees in seconds.
** Average time for one corpus and one tree in seconds.
† Total time for one corpus and all fifty trees in seconds.
‡ Average time for one document and all fifty trees in seconds.

Table 7 shows the phrases selected by the baseline configuration of C4.5 for three articles from the journal article testing subset. In these three examples, the desired number of phrases is set to nine. The phrases in bold match the author’s phrases, according to the Iterated Lovins stemming algorithm.[11]

Experiment 1B tests the hypothesis that bagging improves the performance of C4.5 on the task of automatic keyphrase extraction. Figure 2 shows the precision when the desired number of phrases is set to 5, 7, 9, 11, 13, and 15. The number of trees is set to 1, 25, and 50.

10. The decision tree routines were written by Quinlan (1993). We wrote the feature vector generation routines and the random sampling routines. The code was carefully written for speed. All of the code was written in C and executed on a Pentium II 233 running Windows NT 4.0.

11. The duplication of “probability” and “probabilities” in C4.5’s keyphrases for the first document is due to the limitations of the Iterated Lovins stemming algorithm (see Table 1). Usually duplication of this kind is eliminated during the formation of the feature vectors (see Section 5), since there is only one vector for each unique stemmed phrase.

Table 7: Experiment 1A: Examples of the selected phrases for three articles.

| | |
|---|---|
| Title: | "The Base Rate Fallacy Myth" |
| Author's Keyphrases: | base rate fallacy, Bayes' theorem, decision making, ecological validity, ethics, fallacy, judgment, probability. |
| C4.5's Top Nine Keyphrases: | **judgments**, base rates, **base rate fallacy**, **decision making**, posteriors, **fallacy**, **probability**, rate fallacy, probabilities. |
| Precision: | 0.556 |
| Title: | "Brain Rhythms, Cell Assemblies and Cognition: Evidence from the Processing of Words and Pseudowords" |
| Author's Keyphrases: | brain theory, cell assembly, cognition, event related potentials, ERP, electroencephalograph, EEG, gamma band, Hebb, language, lexical processing, magnetoencephalography, MEG, psychophysiology, periodicity, power spectral analysis, synchrony. |
| C4.5's Top Nine Keyphrases: | cell assemblies, **cognitive**, responses, assemblies, cognitive processing, brain functions, word processing, oscillations, cell. |
| Precision: | 0.111 |
| Title: | "On the Evolution of Consciousness and Language" |
| Author's Keyphrases: | consciousness, language, plans, motivation, evolution, motor system. |
| C4.5's Top Nine Keyphrases: | psychology, **language**, **consciousness**, behavior, **evolution**, cognitive psychology, Bridgeman, organization, modern cognitive psychology. |
| Precision: | 0.333 |

For four of the corpora, the precision tends to rise as the number of trees increases. The exception is the FIPS corpus. As we noted for Experiment 1A, C4.5 has difficulty in generalizing from the journal article training data to the FIPS testing data.

In Table 8, we test the significance of this rising trend, using a paired t-test (Fraser, 1976; Feelders and Verkooijen, 1995). The table shows that, when we look at the five testing collections together, 50 trees are significantly more precise than 1 tree, when the desired number of phrases is set to 15. The only case in which 50 trees are significantly worse than 1 tree is with the FIPS collection.

Experiment 1C tests the hypothesis that stratified sampling (Deming, 1978; Buntine, 1989; Catlett, 1991; Kubat *et al.*, 1998) can help C4.5 handle the skewed class distribution. Figure 3 shows the precision when the percentage of examples in class 1 (positive examples of keyphrases) is set to 1%, 25%, and 50%. For at least three of the corpora, precision tends to fall as the percentage increases.

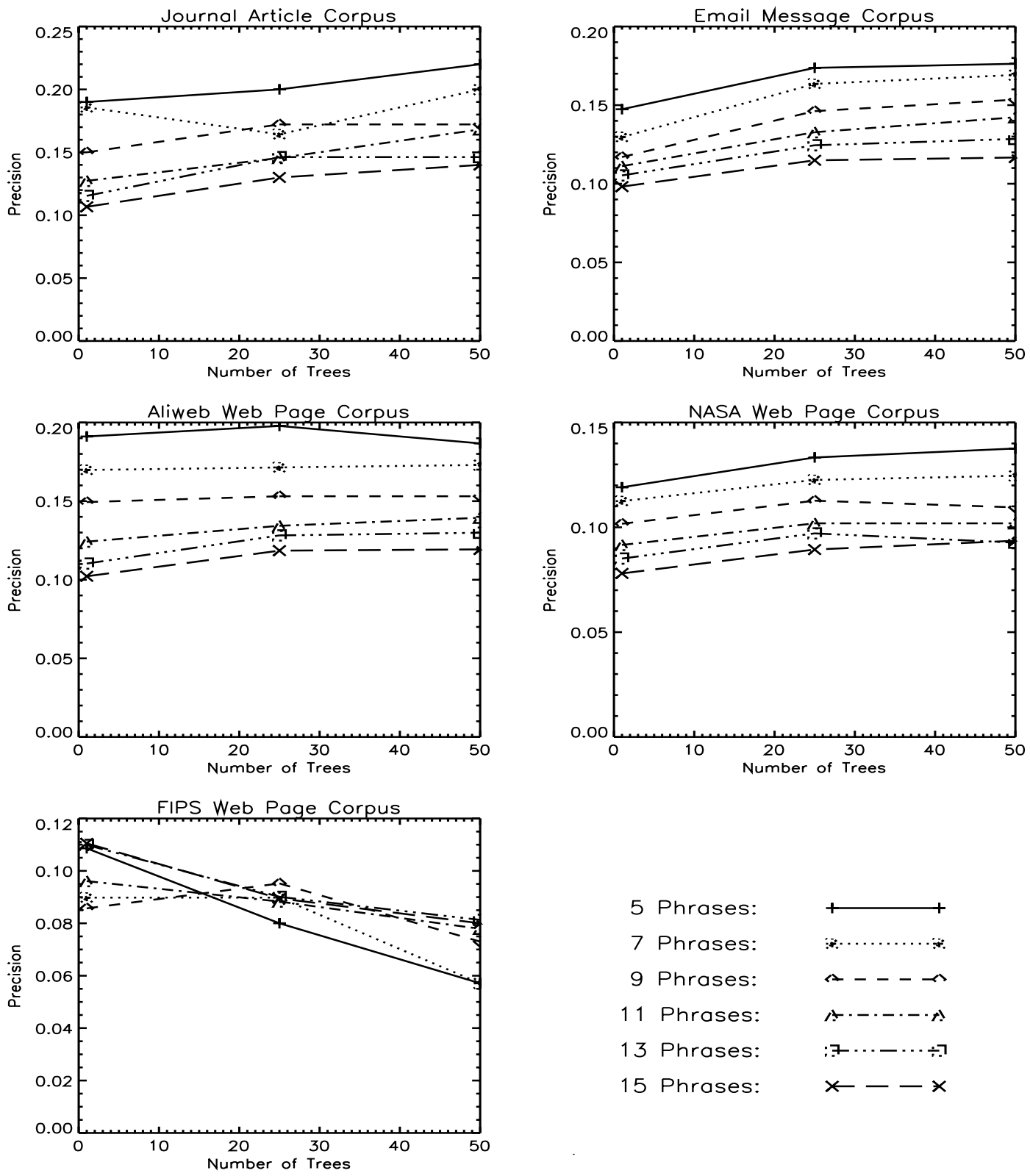

Figure 2: Experiment 1B: The effect of varying the number of trees on precision.

Table 8: Experiment 1B: A comparison of 50 trees with 1 tree.

| Corpus Name | Number of Documents | Number of Phrases | Average Precision ± Standard Deviation | | | Significant with 95% Confidence |
|---|---|---|---|---|---|---|
| | | | 1 Tree | 50 Trees | 50 - 1 | |
| Journal | 20 | 5 | 0.190 ± 0.229 | 0.220 ± 0.182 | 0.030 ± 0.218 | NO |
| | | 15 | 0.107 ± 0.098 | 0.140 ± 0.078 | 0.033 ± 0.085 | NO |
| Email | 76 | 5 | 0.147 ± 0.151 | 0.176 ± 0.160 | 0.029 ± 0.141 | NO |
| | | 15 | 0.098 ± 0.079 | 0.117 ± 0.099 | 0.018 ± 0.060 | YES |
| Aliweb | 90 | 5 | 0.191 ± 0.182 | 0.187 ± 0.166 | -0.004 ± 0.164 | NO |
| | | 15 | 0.102 ± 0.076 | 0.119 ± 0.082 | 0.017 ± 0.057 | YES |
| NASA | 141 | 5 | 0.119 ± 0.137 | 0.138 ± 0.129 | 0.018 ± 0.133 | NO |
| | | 15 | 0.078 ± 0.066 | 0.094 ± 0.069 | 0.016 ± 0.048 | YES |
| FIPS | 35 | 5 | 0.109 ± 0.101 | 0.057 ± 0.092 | -0.051 ± 0.148 | YES |
| | | 15 | 0.111 ± 0.067 | 0.080 ± 0.062 | -0.031 ± 0.078 | YES |
| All | 362 | 5 | 0.146 ± 0.158 | 0.155 ± 0.151 | 0.009 ± 0.151 | NO |
| | | 15 | 0.093 ± 0.074 | 0.106 ± 0.080 | 0.013 ± 0.060 | YES |

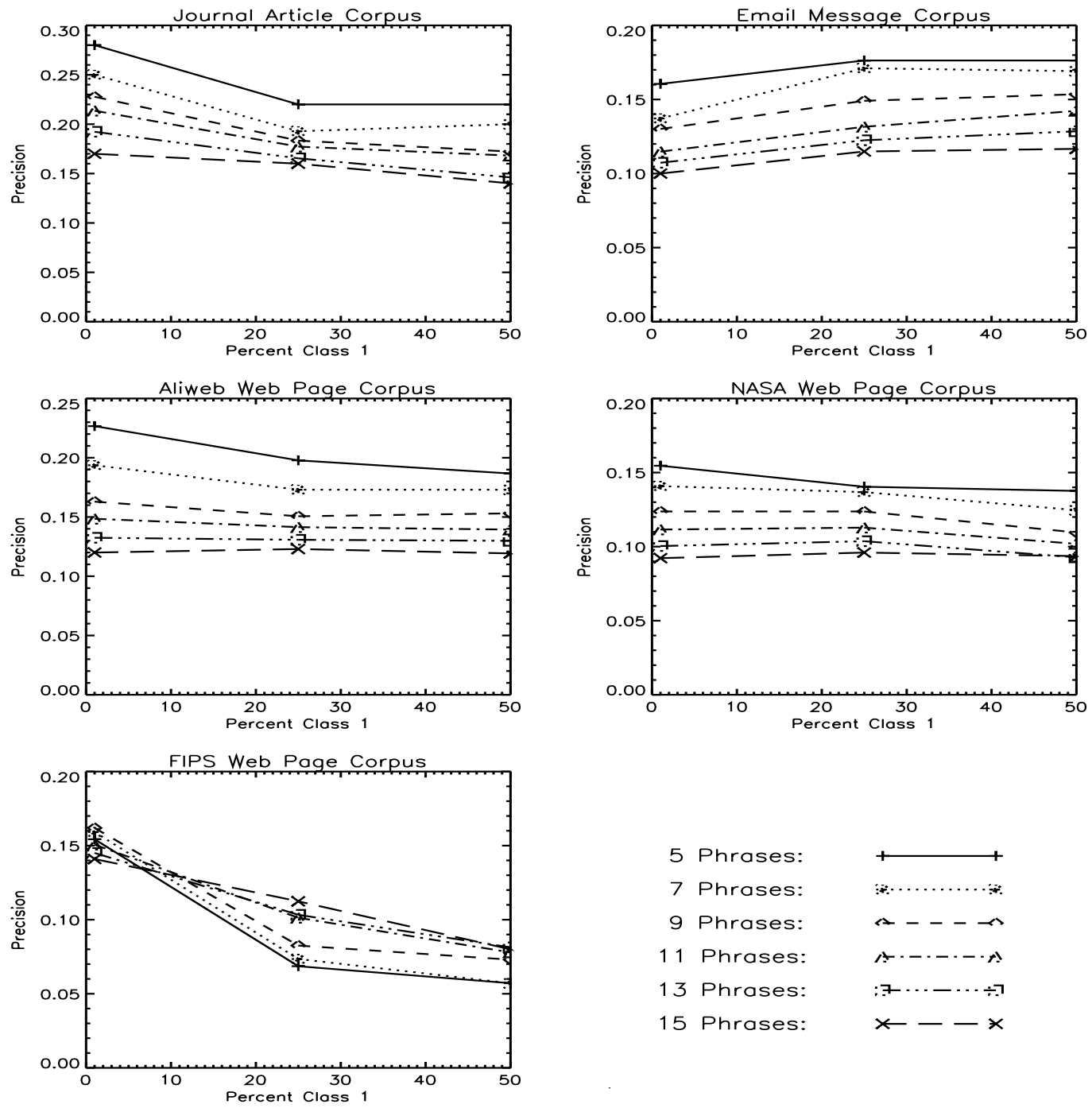

Figure 3: Experiment 1C: The effect of varying the percentage of class 1 on precision.

Table 9: Experiment 1C: A comparison of 1% positive examples with 50% positive examples.

| Corpus Name | Number of Documents | Number of Phrases | Average Precision ± Standard Deviation | | | Significant with 95% Confidence |
|---|---|---|---|---|---|---|
| | | | 1% Class 1 | 50% Class 1 | 50 - 1 | |
| Journal | 20 | 5 | 0.280 ± 0.255 | 0.220 ± 0.182 | -0.060 ± 0.216 | NO |
| | | 15 | 0.170 ± 0.113 | 0.140 ± 0.078 | -0.030 ± 0.103 | NO |
| Email | 76 | 5 | 0.161 ± 0.160 | 0.176 ± 0.160 | 0.016 ± 0.145 | NO |
| | | 15 | 0.100 ± 0.081 | 0.117 ± 0.099 | 0.017 ± 0.055 | YES |
| Aliweb | 90 | 5 | 0.227 ± 0.190 | 0.187 ± 0.166 | -0.040 ± 0.135 | YES |
| | | 15 | 0.120 ± 0.074 | 0.119 ± 0.082 | -0.001 ± 0.048 | NO |
| NASA | 141 | 5 | 0.155 ± 0.159 | 0.138 ± 0.129 | -0.017 ± 0.138 | NO |
| | | 15 | 0.092 ± 0.068 | 0.094 ± 0.069 | 0.001 ± 0.045 | NO |
| FIPS | 35 | 5 | 0.154 ± 0.162 | 0.057 ± 0.092 | -0.097 ± 0.184 | YES |
| | | 15 | 0.141 ± 0.066 | 0.080 ± 0.062 | -0.061 ± 0.063 | YES |
| All | 362 | 5 | 0.181 ± 0.177 | 0.155 ± 0.151 | -0.026 ± 0.151 | YES |
| | | 15 | 0.110 ± 0.078 | 0.106 ± 0.080 | -0.004 ± 0.058 | NO |

Table 9 shows that, when we look at the five testing collections together, there is a significant drop in precision when 50% of the samples are positive examples, compared to 1%. Only the email collection appears to benefit from balanced sampling of the classes. We must reject the hypothesis that stratified sampling (Deming, 1978; Buntine, 1989; Catlett, 1991; Kubat *et al.*, 1998) is useful for our data. Although our preliminary experiments with the training data suggested that stratified sampling would be beneficial, the hypothesis is not supported by the testing data.

Experiment 1D tests the hypothesis that sampling 1% of the training data results in better precision than larger samples. Figure 4 shows the precision when the sample size is 1%, 25%, and 50%. For three of the copora, increasing the sample size tends to decrease the precision. The exceptions are the email message corpus and the FIPS web page corpus.

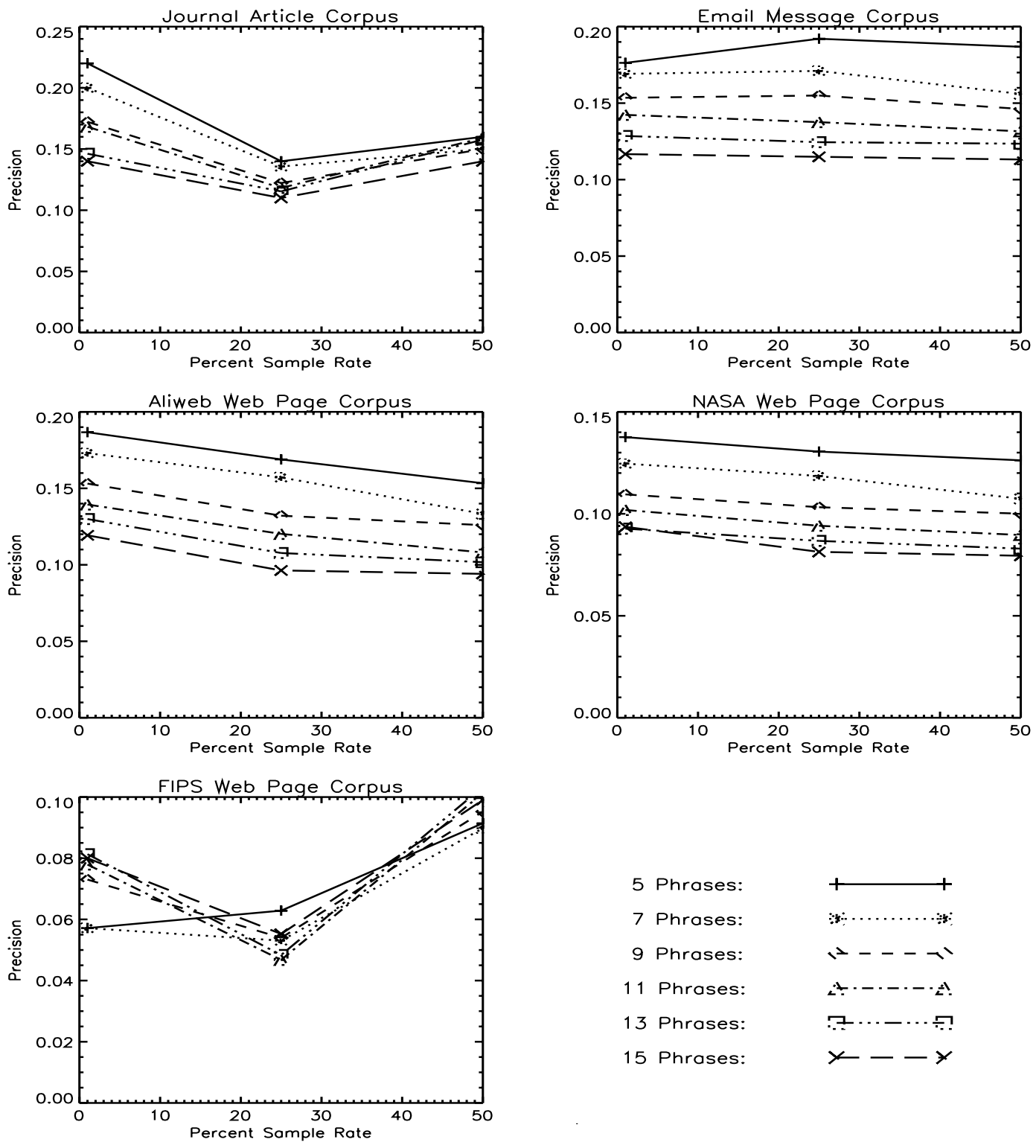

Figure 4: Experiment 1D: The effect of varying the sample rate on precision.

In Table 10, we test the significance of this trend, using a paired t-test. The table shows that, when we look at the five testing collections together, a 1% sample rate yields better precision than a 50% sample rate, when the desired number of phrases is set to 15. This supports the hypothesis that a relatively small sample size is better for bagging than a large sample. This is expected, since bagging works best when the combined models are heterogeneous (Breiman, 1996a, 1996b; Quinlan, 1996). Increasing the sample size tends to make the models more homogenous.

Table 10: Experiment 1D: A comparison of 1% sample rate with 50% sample rate.

| Corpus Name | Number of Documents | Number of Phrases | Average Precision ± Standard Deviation | | | Significant with 95% Confidence |
|---|---|---|---|---|---|---|
| | | | 1% Sample Rate | 50% Sample Rate | 50 - 1 | |
| Journal | 20 | 5 | 0.220 ± 0.182 | 0.160 ± 0.139 | -0.060 ± 0.131 | NO |
| | | 15 | 0.140 ± 0.078 | 0.140 ± 0.094 | 0.000 ± 0.061 | NO |
| Email | 76 | 5 | 0.176 ± 0.160 | 0.187 ± 0.161 | 0.011 ± 0.146 | NO |
| | | 15 | 0.117 ± 0.099 | 0.113 ± 0.102 | -0.004 ± 0.062 | NO |
| Aliweb | 90 | 5 | 0.187 ± 0.166 | 0.153 ± 0.144 | -0.033 ± 0.168 | NO |
| | | 15 | 0.119 ± 0.082 | 0.094 ± 0.071 | -0.025 ± 0.064 | YES |
| NASA | 141 | 5 | 0.138 ± 0.129 | 0.126 ± 0.136 | -0.011 ± 0.143 | NO |
| | | 15 | 0.094 ± 0.069 | 0.079 ± 0.057 | -0.014 ± 0.055 | YES |
| FIPS | 35 | 5 | 0.057 ± 0.092 | 0.091 ± 0.112 | 0.034 ± 0.133 | NO |
| | | 15 | 0.080 ± 0.062 | 0.099 ± 0.059 | 0.019 ± 0.050 | YES |
| All | 362 | 5 | 0.155 ± 0.151 | 0.144 ± 0.144 | -0.010 ± 0.150 | NO |
| | | 15 | 0.106 ± 0.080 | 0.095 ± 0.076 | -0.011 ± 0.060 | YES |

## 7. GenEx: A Hybrid Genetic Algorithm for Keyphrase Extraction

We have experimented with many ways of applying C4.5 to automatic keyphrase extraction. The preceding section presented a few of these experiments. During the course of our experimentation, we came to believe that a tailor-made algorithm for learning to extract keyphrases might be able to achieve better precision than a general-purpose learning algorithm such as C4.5. As we discuss in Section 10, the main insight that we gained from working with C4.5 was that we might achieve better performance in this domain if we could find a way to embed *specialized procedural domain knowledge* into our keyphrase extraction algorithm. This motivated us to develop the *GenEx* algorithm.

GenEx has two components, the *Genitor* genetic algorithm (Whitley, 1989) and the *Extractor* keyphrase extraction algorithm (Turney, 1997, 1999). Extractor takes a document as input and produces a list of keyphrases as output. Extractor has twelve parameters that determine how it processes the input text. In GenEx, the parameters of Extractor are tuned by the Genitor genetic algorithm (Whitley, 1989), to maximize performance (fitness) on training data. Genitor is used to tune Extractor, but Genitor is no longer needed once the training process is complete. When we know the best parameter values, we can discard Genitor. Thus the learning system is called GenEx (Genitor plus Extractor) and the trained system is called Extractor (GenEx minus Genitor).

### 7.1 Extractor

What follows is a conceptual description of the Extractor algorithm. For clarity, we describe Extractor at an abstract level that ignores efficiency considerations. That is, the actual Extractor software is essentially an efficient implementation of the following algorithm.[12] In the following, the twelve parameters appear in small capitals (see Table 11 for a list of the parameters). There are ten steps to the Extractor algorithm:

1. **Find Single Stems:** Make a list of all of the words in the input text. Drop words with less than three characters. Drop stop words, using a given stop word list. Convert all remaining words to lower case. Stem the words by truncating them at STEM_LENGTH characters.

The advantages of this simple form of stemming (stemming by truncation) are speed and flexibility. Stemming by truncation is much faster than either the Lovins (1968) or Porter (1980) stemming algorithms. The aggressiveness of the stemming can be adjusted by changing STEM_LENGTH. This gives Genitor control over the level of aggressiveness.

2. **Score Single Stems:** For each unique stem, count how often the stem appears in the text and note when it first appears. If the stem “evolut” first appears in the word “Evolution”, and “Evolution” first appears as the tenth word in the text, then the first appearance of “evolut” is said to be in position 10. Assign a score to each stem. The score is the number of times the stem appears in the text, multiplied by a factor. If the stem first appears

12. Extractor is written in C. A demonstration version of Extractor is available at http://extractor.iit.nrc.ca/. The demonstration version has been trained already; it does not allow the user to make any adjustments.

before FIRST_LOW_THRESH, then multiply the frequency by FIRST_LOW_FACTOR. If the stem first appears after FIRST_HIGH_THRESH, then multiply the frequency by FIRST_HIGH_FACTOR.

Typically FIRST_LOW_FACTOR is greater than one and FIRST_HIGH_FACTOR is less than one. Thus, early, frequent stems receive a high score and late, rare stems receive a low score. This gives Genitor control over the weight of early occurrence versus the weight of frequency.

3. **Select Top Single Stems:** Rank the stems in order of decreasing score and make a list of the top NUM_WORKING single stems.

Cutting the list at NUM_WORKING, as opposed to allowing the list to have an arbitrary length, improves the efficiency of Extractor. It also acts as a filter for eliminating lower quality stems.

4. **Find Stem Phrases:** Make a list of all phrases in the input text. A phrase is defined as a sequence of one, two, or three words that appear consecutively in the text, with no intervening stop words or punctuation. Stem each phrase by truncating each word in the phrase at STEM_LENGTH characters.

In our corpora, phrases of four or more words are relatively rare. Therefore Extractor only considers phrases of one, two, or three words.

5. **Score Stem Phrases:** For each stem phrase, count how often the stem phrase appears in the text and note when it first appears. Assign a score to each phrase, exactly as in step 2, using the parameters FIRST_LOW_FACTOR, FIRST_LOW_THRESH, FIRST_HIGH_FACTOR, and FIRST_HIGH_THRESH. Then make an adjustment to each score, based on the number of stems in the phrase. If there is only one stem in the phrase, do nothing. If there are two stems in the phrase, multiply the score by FACTOR_TWO_ONE. If there are three stems in the phrase, multiply the score by FACTOR_THREE_ONE.

Typically FACTOR_TWO_ONE and FACTOR_THREE_ONE are greater than one, so this adjust-

ment will increase the score of longer phrases. A phrase of two or three stems is necessarily never more frequent than the most frequent single stem contained in the phrase. The factors FACTOR_TWO_ONE and FACTOR_THREE_ONE are designed to boost the score of longer phrases, to compensate for the fact that longer phrases are expected to otherwise have lower scores than shorter phrases.

6. **Expand Single Stems:** For each stem in the list of the top NUM_WORKING single stems, find the highest scoring stem phrase of one, two, or three stems that contains the given single stem. The result is a list of NUM_WORKING stem phrases. Keep this list ordered by the scores calculated in step 2.

Now that the single stems have been expanded to stem phrases, we no longer need the scores that were calculated in step 5. That is, the score for a stem phrase (step 5) is now replaced by the score for its corresponding single stem (step 2). The reason is that the adjustments to the score that were introduced in step 5 are useful for expanding the single stems to stem phrases, but they are not useful for comparing or ranking stem phrases.

7. **Drop Duplicates:** The list of the top NUM_WORKING stem phrases may contain duplicates. For example, two single stems may expand to the same two-word stem phrase. Delete duplicates from the ranked list of NUM_WORKING stem phrases, preserving the highest ranked phrase.

For example, suppose that the stem “evolu” (e.g., “evolution” truncated at five characters) appears in the fifth position in the list of the top NUM_WORKING single stems and “psych” (e.g., “psychology” truncated at five characters) appears in the tenth position. When the single stems are expanded to stem phrases, we might find that “evolu psych” (e.g., “evolutionary psychology” truncated at five characters) appears in the fifth and tenth positions in the list of stem phrases. In this case, we delete the phrase in the tenth position. If there are duplicates, then the list now has fewer than NUM_WORKING stem phrases.

8. **Add Suffixes:** For each of the remaining stem phrases, find the most frequent corre-

sponding whole phrase in the input text. For example, if “evolutionary psychology” appears ten times in the text and “evolutionary psychologist” appears three times, then “evolutionary psychology” is the more frequent corresponding whole phrase for the stem phrase “evolu psych”. When counting the frequency of whole phrases, if a phrase has an ending that indicates a possible adjective, then the frequency for that whole phrase is set to zero. An ending such as “al”, “ic”, “ible”, etc., indicates a possible adjective. Adjectives in the middle of a phrase (for example, the second word in a three-word phrase) are acceptable; only phrases that end in adjectives are penalized. Also, if a phrase contains a verb, the frequency for that phrase is set to zero. To check for verbs, we use a list of common verbs. A word that might be either a noun or a verb is included in this list only when it is much more common for the word to appear as a verb than as a noun.

For example, suppose the input text contains “manage”, “managerial”, and “management”. If STEM_LENGTH is, say, five, the stem “manag” will be expanded to “management” (a noun), because the frequency of “managerial” will be set to zero (because it is an adjective, ending in “al”) and the frequency of “manage” will be set to zero (because it is a verb, appearing in the list of common verbs). Although “manage” and “managerial” would not be output, their presence in the input text helps to boost the score of the stem “manag” (as measured in step 2), and thereby increase the likelihood that “management” will be output.

9. **Add Capitals:** For each of the whole phrases (phrases with suffixes added), find the best capitalization, where *best* is defined as follows. For each word in a phrase, find the capitalization with the least number of capitals. For a one-word phrase, this is the best capitalization. For a two-word or three-word phrase, this is the best capitalization, unless the capitalization is inconsistent. The capitalization is said to be inconsistent when one of the words has the capitalization pattern of a proper noun but another of the words does not appear to be a proper noun (e.g., “Turing test”). When the capitalization is inconsistent, see whether it can be made consistent by using the capitalization with the second lowest number of capitals (e.g., “Turing Test”). If it cannot be made consistent, use the inconsistent capitalization. If it can be made consistent, use the consistent capitalization.

For example, given the phrase “psychological association”, the word “association” might appear in the text only as “Association”, whereas the word “psychological” might appear in the text as “PSYCHOLOGICAL”, “Psychological”, and “psychological”. Using the least number of capitals, we get “psychological Association”, which is inconsistent. However, it can be made consistent, as “Psychological Association”.

10. **Final Output:** We now have an ordered list of mixed-case (upper and lower case, if appropriate) phrases with suffixes added. The list is ordered by the scores calculated in step 2. That is, the score of each whole phrase is based on the score of the highest scoring single stem that appears in the phrase. The length of the list is at most NUM_WORKING, and is likely less, due to step 7. We now form the final output list, which will have at most NUM_PHRASES phrases. We go through the list of phrases in order, starting with the top-ranked phrase, and output each phrase that passes the following tests, until either NUM_PHRASES phrases have been output or we reach the end of the list. The tests are (1) the phrase should not have the capitalization of a proper noun, unless the flag SUPPRESS_PROPER is set to 0 (if 0 then allow proper nouns; if 1 then suppress proper nouns); (2) the phrase should not have an ending that indicates a possible adjective; (3) the phrase should be longer than MIN_LENGTH_LOW_RANK, where the length is measured by the ratio of the number of characters in the candidate phrase to the number of characters in the average phrase, where the average is calculated for all phrases in the input text that consist of one to three consecutive non-stop words; (4) if the phrase is shorter than MIN_LENGTH_LOW_RANK, it may still be acceptable, if its rank in the list of candidate phrases is better than (closer to the top of the list than) MIN_RANK_LOW_LENGTH; (5) if the phrase fails both tests (3) and (4), it may still be acceptable, if its capitalization pattern indicates that it is probably an abbreviation; (6) the phrase should not contain any words that are most commonly used as verbs; (7) the phrase should not match any phrases in a given list of stop phrases (where “match” means equal strings, ignoring case, but including suffixes).

That is, a phrase must pass tests (1), (2), (6), (7), and at least one of tests (3), (4), and (5).

Although our experimental procedure does not consider capitalization or suffixes when comparing machine-generated keyphrases to human-generated keyphrases, steps 8 and 9 are still useful, because some of the screening tests in step 10 are based on capitalization and suffixes. Of course, steps 8 and 9 are essential when the output is for human readers.

### 7.2 Genitor

A genetic algorithm may be viewed as a method for optimizing a string of bits, using techniques that are inspired by biological evolution. A genetic algorithm works with a set of bit strings, called a *population* of *individuals*. The initial population is usually randomly generated. New individuals (new bit strings) are created by randomly changing existing individuals (this operation is called *mutation*) and by combining substrings from *parents* to make new *children* (this operation is called *crossover*). Each individual is assigned a score (called its *fitness*) based on some measure of the quality of the bit string, with respect to a given task. Fitter individuals get to have more children than less fit individuals. As the genetic algorithm runs, new individuals tend to be increasingly fit, up to some asymptote.

Genitor is a *steady-state* genetic algorithm (Whitley, 1989), in contrast to many other genetic algorithms, such as Genesis (Grefenstette 1983, 1986), which are *generational*.[13] A generational genetic algorithm updates its entire population in one batch, resulting in a sequence of distinct generations. A steady-state genetic algorithm updates its population one individual at a time, resulting in a continuously changing population, with no distinct generations. Typically a new individual replaces the least fit individual in the current population.

13. The source code for Genitor (written in C) is available at ftp://ftp.cs.colostate.edu/pub/GENITOR.tar.

Whitley (1989) suggests that steady-state genetic algorithms tend to be more aggressive (they have greater selective pressure) than generational genetic algorithms.

### 7.3 GenEx

The parameters in Extractor are set using the standard machine learning paradigm of supervised learning. The algorithm is tuned with a dataset, consisting of documents paired with target lists of keyphrases. The dataset is divided into training and testing subsets. The learning process involves adjusting the parameters to maximize the match between the output of Extractor and the target keyphrase lists, using the training data. The success of the learning process is measured by examining the match using the testing data.

We assume that the user sets the value of NUM_PHRASES, the desired number of phrases, to a value between five and fifteen. We then set NUM_WORKING to $5 \cdot$ NUM_PHRASES. The remaining ten parameters are set by Genitor. Genitor uses a binary string of 72 bits to represent the ten parameters, as shown in Table 11. We run Genitor with a population size of 50 for 1050 trials (these are default settings). Each trial consists of running Extractor with the parameter settings specified in the given binary string, processing the entire training set. The fitness measure for the binary string is based on the average precision for the whole training set. The final output of Genitor is the highest scoring binary string. Ties are broken by choosing the earlier string.

We first tried to use the average precision on the training set as the fitness measure, but GenEx discovered that it could achieve high average precision by adjusting the parameters so that less than NUM_PHRASES phrases were output. This is not desirable, so we modified the fitness measure to penalize GenEx when less than NUM_PHRASES phrases were output:

$$\text{total_matches} = \text{total number of matches between GenEx and human} \quad (1)$$

$$\text{total_machine_phrases} = \text{total number of phrases output by GenEx} \quad (2)$$

$$\text{precision} = \text{total_matches} / \text{total_machine_phrases} \quad (3)$$

$$\text{num_docs} = \text{number of documents in training set} \quad (4)$$

Table 11: The twelve parameters of Extractor, with types and ranges.

| Parameter Number | Parameter Name | Parameter Type | Parameter Range | Number of Bits |
|---|---|---|---|---|
| 1 | NUM_PHRASES * | integer | [5, 15] | 0 * |
| 2 | NUM_WORKING † | integer | [25, 75] | 0 † |
| 3 | FACTOR_TWO_ONE | real | [1, 3] | 8 |
| 4 | FACTOR_THREE_ONE | real | [1, 5] | 8 |
| 5 | MIN_LENGTH_LOW_RANK | real | [0.3, 3.0] | 8 |
| 6 | MIN_RANK_LOW_LENGTH | integer | [1, 20] | 5 |
| 7 | FIRST_LOW_THRESH | integer | [1, 1000] | 10 |
| 8 | FIRST_HIGH_THRESH | integer | [1, 4000] | 12 |
| 9 | FIRST_LOW_FACTOR | real | [1, 15] | 8 |
| 10 | FIRST_HIGH_FACTOR | real | [0.01, 1.0] | 8 |
| 11 | STEM_LENGTH | integer | [1, 10] | 4 |
| 12 | SUPPRESS_PROPER | boolean | [0, 1] | 1 |
| | Total Number of Bits in Binary String: | | | 72 |

* This parameter is set by the user to the desired value.
† This parameter is set to five times NUM_PHRASES.

$$\text{total_desired} = \text{num_docs} \cdot \text{NUM_PHRASES} \tag{5}$$

$$\text{penalty} = (\text{total_machine_phrases} / \text{total_desired})^2 \tag{6}$$

$$\text{fitness} = \text{precision} \cdot \text{penalty} \tag{7}$$

The penalty factor varies between 0 and 1. It has no effect (i.e., it is 1) when the number of phrases output by GenEx equals the desired number of phrases. The penalty grows (i.e., it approaches 0) with the square of the gap between the desired number of phrases and the actual number of phrases. Preliminary experiments on the training data confirmed that this fitness measure led GenEx to find parameter values with high average precision while ensuring that NUM_PHRASES phrases were output.

Since STEM_LENGTH is modified by Genitor during the GenEx learning process, the fitness measure used by Genitor is not based on stemming by truncation. If the fitness measure were based on stemming by truncation, a change in STEM_LENGTH would change the apparent fitness, even if the actual output keyphrase list remained constant. Therefore fitness is measured with the Iterated Lovins stemmer.

We ran Genitor with a *Selection Bias* of 2.0 and a *Mutation Rate* of 0.2. These are the default settings for Genitor. We used the *Adaptive Mutation* operator and the *Reduced Surrogate Crossover* operator (Whitley, 1989). *Adaptive Mutation* determines the appropriate level of mutation for a child according to the hamming distance between its two parents; the less the difference, the higher the mutation rate. *Reduced Surrogate Crossover* first identifies all positions in which the parent strings differ. Crossover points are only allowed to occur in these positions.

A comparison of Extractor with the feature vectors we used with C4.5 shows that GenEx and C4.5 are learning with essentially the same feature sets. The two algorithms have access to the same information, but they learn different kinds of models of keyphrases. Here are some of the more significant differences between GenEx and C4.5 (as we have used it here):

1. Given a set of phrases with a shared single-word stem (for example, the set of phrases {“learning”, “machine learning”, “learnability”} shares the single-word stem “learn”), GenEx tends to choose the best member of the set, rather than choosing the whole set. GenEx first identifies the shared single-word stem (steps 1 to 3) and then looks for the best representative phrase in the set (steps 4 to 6). C4.5 tends to choose several members from the set, if it chooses any of them.[14]
2. GenEx can adjust the aggressiveness of the stemming, by adjusting STEM_LENGTH. C4.5 must take the stems that are given in the training data.[15]
3. C4.5 is designed to yield high accuracy. GenEx is designed to yield high precision for a given NUM_PHRASES. High precision does not necessarily correspond to high accuracy.
4. C4.5 uses the same model (the same set of decision trees) for all values of

14. It would be possible to add a post-processing step that winnows the phrases selected by C4.5, but this would be a step down the path that leads from the general-purpose C4.5 algorithm to the custom-made GenEx algorithm. See the discussion in Section 10.

15. Again, it would be possible to adjust the stemming procedure externally, by cross-validation, but this is another step down the path that leads from C4.5 to GenEx.

NUM_PHRASES. With C4.5, we select the top NUM_PHRASES most probable feature vectors, but our estimate of probability is not sensitive to the value of NUM_PHRASES. On the other hand, Genitor tunes Extractor differently for each desired value of NUM_PHRASES.

5. GenEx might output less than the desired number of phrases, NUM_PHRASES, but C4.5 (as we use it here) always generates exactly NUM_PHRASES phrases. Therefore, in the following experiments, performance is measured by the average precision, where precision is defined by (8), not by (9). Equation (8) ensures that GenEx cannot spuriously boost its score by generating fewer phrases than the user requests.[16]

$$\text{precision} = \text{number of matches} \,/\, \text{desired number of machine-generated phrases} \tag{8}$$

$$\text{precision} = \text{number of matches} \,/\, \text{actual number of machine-generated phrases} \tag{9}$$

## 8. Experiment 2: Learning to Extract Keyphrases with GenEx

This set of experiments compares GenEx to C4.5. In Figure 5, we compare GenEx to both the baseline configuration of C4.5 (Experiment 1A) and the best configuration of C4.5 (Experiment 1C). It is not quite fair to GenEx to use the best configuration of C4.5, because we only know that it is the best by looking at the testing data, but GenEx does not have the advantage of any information from the testing data. However, the performance of GenEx is good enough that we can afford to be generous to C4.5.

Figure 5 shows the average precision on each testing corpus, with the desired number of phrases set to 5, 7, 9, 11, 13, and 15. In Table 12, we test the significance of the difference between GenEx and the best configuration of C4.5. The table shows that, when we look at the five testing collections together, GenEx is significantly more precise. There is no case in which the performance of GenEx is below the performance of C4.5.

Table 13 shows the training and testing time for GenEx. For ease of comparison, we have

16. If GenEx does generate fewer than NUM_PHRASES phrases, then we could randomly select further phrases until we have NUM_PHRASES phrases. Thus (8) does not misrepresent the performance of GenEx. Note that we do not use the fitness measure (7) to evaluate the performance of GenEx.

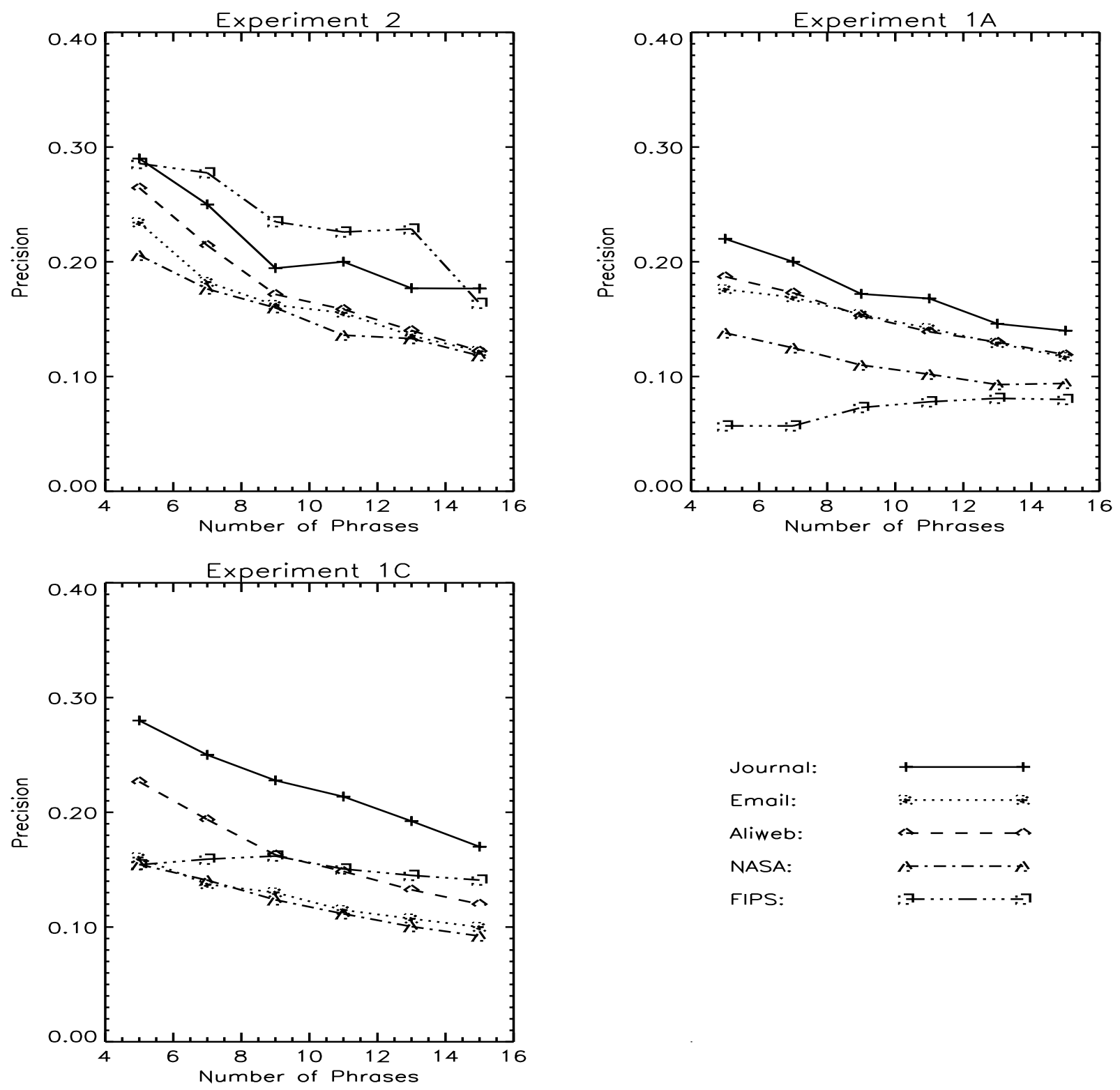

Figure 5: A comparison of GenEx (Experiment 2) with C4.5 (Experiments 1A and 1C).

Table 12: A comparison of GenEx (Experiment 2) with C4.5 (Experiment 1C).

| Corpus Name | Number of Documents | Number of Phrases | Average Precision ± Standard Deviation | | | Significant with 95% Confidence |
|---|---|---|---|---|---|---|
| | | | GenEx | C4.5 (1C) | GenEx - C4.5 | |
| Journal | 20 | 5 | 0.290 ± 0.247 | 0.280 ± 0.255 | 0.010 ± 0.137 | NO |
| | | 15 | 0.177 ± 0.130 | 0.170 ± 0.113 | 0.007 ± 0.061 | NO |
| Email | 76 | 5 | 0.234 ± 0.205 | 0.161 ± 0.160 | 0.074 ± 0.166 | YES |
| | | 15 | 0.122 ± 0.105 | 0.100 ± 0.081 | 0.022 ± 0.073 | YES |
| Aliweb | 90 | 5 | 0.264 ± 0.177 | 0.227 ± 0.190 | 0.038 ± 0.185 | NO |
| | | 15 | 0.122 ± 0.077 | 0.120 ± 0.074 | 0.002 ± 0.077 | NO |
| NASA | 141 | 5 | 0.206 ± 0.172 | 0.155 ± 0.159 | 0.051 ± 0.136 | YES |
| | | 15 | 0.118 ± 0.080 | 0.092 ± 0.068 | 0.026 ± 0.068 | YES |
| FIPS | 35 | 5 | 0.286 ± 0.170 | 0.154 ± 0.162 | 0.131 ± 0.222 | YES |
| | | 15 | 0.164 ± 0.078 | 0.141 ± 0.066 | 0.023 ± 0.081 | NO |
| All | 362 | 5 | 0.239 ± 0.187 | 0.181 ± 0.177 | 0.058 ± 0.167 | YES |
| | | 15 | 0.128 ± 0.089 | 0.110 ± 0.078 | 0.018 ± 0.073 | YES |

reproduced the comparable times from Table 6. Unlike C4.5, GenEx was trained separately for each of the six values of NUM_PHRASES, 5, 7, 9, 11, 13, and 15. In comparison with C4.5,

GenEx is much slower to train, but also much faster after it has been trained. (The same computer was used for timing C4.5 and GenEx.)

Table 13: Training and testing time for GenEx and C4.5.

| Train/Test | Corpus Name | Number of Documents | Total Time in Hours:Minutes:Seconds * | | Average Time in Hours:Minutes:Seconds ** | |
|---|---|---|---|---|---|---|
| | | | GenEx | C4.5 | GenEx | C4.5 |
| Training | Journal | 55 | 48:28:03 | 00:04:10 | 08:04:40 | 00:04:10 |
| | Email | 235 | 14:54:15 | 00:01:29 | 02:29:02 | 00:01:29 |
| | | | Total Time in Seconds † | | Average Time in Seconds ‡ | |
| | | | GenEx | C4.5 | GenEx | C4.5 |
| Testing | Journal | 20 | 5 | 37 | 0.25 | 1.9 |
| | Email | 76 | 4 | 106 | 0.05 | 1.4 |
| | Aliweb | 90 | 6 | 130 | 0.07 | 1.4 |
| | NASA | 141 | 8 | 200 | 0.06 | 1.4 |
| | FIPS | 35 | 12 | 86 | 0.34 | 2.5 |

* Total time for one corpus and all six values of NUM_PHRASES in hours:minutes:seconds.
** Average time for one corpus and one value of NUM_PHRASES in hours:minutes:seconds.
† Total time for one corpus and one value of NUM_PHRASES in seconds.
‡ Average time for one document and one value of NUM_PHRASES in seconds.

Table 14 presents some examples of the phrases selected by GenEx, when NUM_PHRASES is set to nine. Matches with the authors (according to the Iterated Lovins stemming algorithm) are in bold.

## 9. Experiment 3: Human Evaluation of GenEx Keyphrases

It is not obvious whether a precision of, say, 29% for five phrases is good or bad. We believe that it is useful to know that one algorithm has a precision of 29% (for a given corpus and a given desired number of phrases) while another algorithm has a precision of 15% (for the same corpus and the same number of phrases), but a precision of 29% has no significance by itself. What we would really like to know is, what percentage of the keyphrases generated by GenEx are acceptable to a human reader? The following experiment suggests that the answer to this question is that, on average, about 80% of the keyphrases extracted by GenEx are acceptable to a human reader.

Table 14: Experiment 2: Examples of the selected phrases for three articles from *Psycoloquy*.

| | |
|---|---|
| Title: | "The Base Rate Fallacy Myth" |
| Author's Keyphrases: | base rate fallacy, Bayes' theorem, decision making, ecological validity, ethics, fallacy, judgment, probability. |
| GenEx's Top Nine Keyphrases: | base rates, **judgments**, **probability**, decision, **base rate fallacy**, prior, experiments, **decision making**, probabilistic information. |
| Precision: | 0.444 |
| Title: | "Brain Rhythms, Cell Assemblies and Cognition: Evidence from the Processing of Words and Pseudowords" |
| Author's Keyphrases: | brain theory, cell assembly, cognition, event related potentials, ERP, electroencephalograph, EEG, gamma band, Hebb, language, lexical processing, magnetoencephalography, MEG, psychophysiology, periodicity, power spectral analysis, synchrony. |
| GenEx's Top Nine Keyphrases: | neurons, pseudowords, responses, cell assemblies, ignition, activation, brain, cognitive processing, gamma-band responses. |
| Precision: | 0.000 |
| Title: | "On the Evolution of Consciousness and Language" |
| Author's Keyphrases: | consciousness, language, plans, motivation, evolution, motor system. |
| GenEx's Top Nine Keyphrases: | **plans**, **consciousness**, **language**, planning, psychology, behavior, memory, cognitive psychology, plan-executing. |
| Precision: | 0.333 |

We created an on-line demonstration of GenEx on the web. The demonstration allows the user to enter any URL for processing. The software then downloads the HTML at the given URL and sends it to Extractor. Extractor uses the parameter values that were learned from the training sets discussed here. For web pages with less than 3000 words, Extractor used the parameters learned from the Email Messages corpus. For web pages with 3000 words or more, Extractor used the parameters learned from the Journal Article corpus. The keyphrases are shown to the user, who may then rate each keyphrase as "good" or "bad". Some or all keyphrases may be left unrated (we call this "no opinion"). The number of keyphrases is fixed at seven, to keep the interface simple. Part of the interface is shown in Figure 6.

Over a seven-month period, visitors to our web site rated keyphrases generated from 385 different web pages. For this experiment, we deleted from our log of ratings all ratings by NRC employees (who may be biased towards Extractor) and all ratings of web pages in languages other than English. (Extractor 5.0, the current version at the time of writing, works with documents in English, French, German, and Japanese.) We were left with evaluations for keyphrases from 267 web pages, as summarized in Table 15.

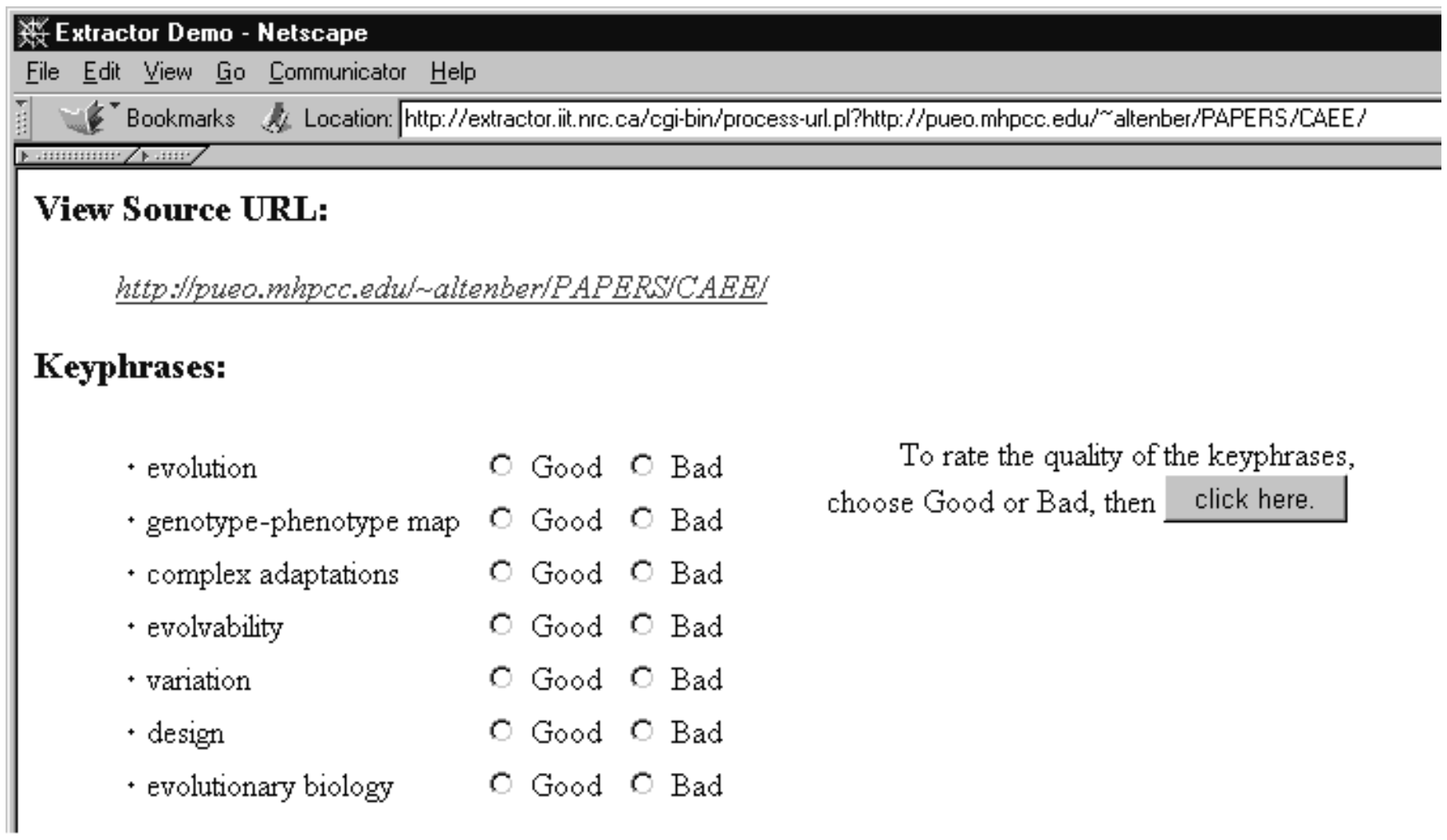

Figure 6: This is an illustration of the on-line interface to Extractor. In this example, the user has entered a URL and the resulting keyphrases are displayed. The user may rate the quality of the keyphrases as "good" or "bad". It is not necessary to rate all or any of the keyphrases.

Table 15: Human evaluation of automatically generated keyphrases for web pages.

| | | |
|---|---|---|
| Number of Voters: | 205 | |
| Number of Documents: | 267 | |
| Number of Keyphrases: | 1869 | |
| Maximum Documents per Voter: | 5 | |
| Good: | 1159 | 62.0% |
| Bad: | 339 | 18.1% |
| No Opinion: | 371 | 19.9% |
| Acceptable (Good + No Opinion): | 1530 | 81.9% |

We interpret the data as supporting the hypothesis that about 80% of the keyphrases are acceptable, on average (*acceptable* meaning *not bad*). The voters were anonymous (we have only their IP addresses) and self-selected. There was no reward for voting and no reason for giving false information. The only instructions to the voters were the words that are displayed in Figure 6.

## 10. Discussion

We have presented two approaches to the task of learning to extract keyphrases from text. The first approach was to apply the C4.5 decision tree induction algorithm (Quinlan, 1993), using soft-thresholds (Carter and Catlett, 1987; Quinlan, 1987, 1990, 1993), bagging (Breiman, 1996a, 1996b; Quinlan, 1996) and stratified sampling (Deming, 1978; Buntine, 1989; Catlett, 1991; Kubat *et al.*, 1998). The experiments support the claim that bagging is helpful for this task, but stratified sampling is not helpful.

Our experience with C4.5 led us to suspect that a custom-designed algorithm might perform better than the general-purpose C4.5 algorithm. Our second approach was to use the Genitor genetic algorithm (Whitley, 1989) to tune the parameters in a special-purpose, parameterized keyphrase extraction algorithm (Extractor). We presented the GenEx algorithm and experiments that support the claim that GenEx performs better than C4.5.

C4.5 has become a standard benchmark in machine learning research for evaluating the performance of supervised classification algorithms. We believe that a thorough effort to apply C4.5, using nine different configurations, is more significant and interesting than a cursory examination of several alternative machine learning algorithms.

One way to contrast C4.5 (as we used it here) and GenEx is to compare how they use domain knowledge. There are two types of domain knowledge implicit in our use of C4.5. The design of the feature vectors (Table 4) clearly involves a significant amount of domain knowledge. The use of soft-thresholds, bagging, and stratified sampling incorporates another kind of domain knowledge. We can characterize the first type of domain knowledge as *declarative* and the second type as *procedural*. Similarly, the GenEx algorithm embodies both declarative and procedural domain knowledge. C4.5 and GenEx embody essentially the same declarative knowledge, about the kinds of features that are relevant for detecting keyphrases, but GenEx contains much more specialized and detailed procedural domain knowledge than C4.5. Soft-thresholds, bagging, and stratified sampling are useful for many tasks besides keyphrase extraction, but the ten steps of the Extractor algorithm are intended specifically for keyphrase extraction. In our application of C4.5, we avoided using any specialized procedural domain knowledge. The experimental results support the view that specialized procedural domain knowledge is valuable for learning to extract keyphrases from text.

As we mentioned earlier (in Section 3), in the time since we submitted this paper, Frank *et al.* (1999) developed *Kea*, which takes a Bayesian approach to keyphrase extraction. Kea includes procedural domain knowledge, in the form of a final post-processing operation. A candidate keyphrase is eliminated if it is contained within another candidate keyphrase that has a higher score. Frank *et al.* (1999) compare GenEx, Kea, and C4.5, when both Kea and C4.5 include their post-processing operation. On their corpora (a subset of our corpora), they found no statistically significant differences among the three algorithms. These results also support the importance of specialized procedural domain knowledge. The results suggest that the amount of procedural domain knowledge that is required for good performance on this task may be smaller than we suspected.

We believe that precision (the percentage of the machine-generated keyphrases that match the human-generated keyphrases) is a good measure for comparing keyphrase extraction algorithms (assuming they are compared using the same documents and extracting the same number of keyphrases), but precision does not capture the subjective quality of the keyphrases. Although our precision never went above 30%, human evaluation of the quality of the machine-generated keyphrases suggests that about 80% of the keyphrases are of acceptable quality.

## 11. Future Work and Limitations

A limitation of our approach is that authors often use synonyms, to avoid boring the reader with repetition. Unfortunately, repetition is a major clue for GenEx and C4.5 (using the features in Table 4) that a candidate phrase is a keyphrase. We believe that the results could be significantly improved by adding some kind of synonym detection to the keyphrase extrac-tion algorithms.

Extractor 5.0 (the current version at the time of writing) works with documents in English, French, German, and Japanese (Mathieu, 1999). We are currently working on adding Spanish and Korean. We are also working on adding sentence extraction capability, in addition to the existing phrase extraction capability (Luhn, 1958; Edmundson, 1969; Marsh *et al.*, 1984; Paice, 1990; Paice and Jones, 1993; Johnson *et al.*, 1993; Salton *et al.*, 1994; Kupiec *et al.*, 1995; Brandow *et al.*, 1995; Jang and Myaeng, 1997).

## 12. Conclusion

The experimental results support the claim that specialized procedural domain knowledge is valuable for learning to extract keyphrases from text. A keyphrase extraction algorithm incorporating such specialized knowledge (GenEx) performed significantly better than an algorithm without such knowledge (C4.5).

Extractor can make a keyphrase list for an average journal article in one quarter of a second (Table 13). The speed of Extractor makes it possible to use keyphrases in applications where it would not be economically feasible to use human-generated keyphrases. Subjective human evaluation of the keyphrases generated by Extractor suggests that about 80% of the keyphrases are acceptable to human readers. This level of performance should be satisfactory for a wide variety of applications.

## Acknowledgments

Thanks to the anonymous referees of *Information Retrieval* for their very helpful comments on this paper. Thanks to Elaine Sin of the University of Calgary for creating the keyphrases for the email message corpus. Thanks to Jérôme Mathieu of Tetranet Software (previously of NRC) for suggesting the user feedback mechanism illustrated in Figure 6. Thanks to my colleagues at NRC and the University of Ottawa for their support, encouragement, and constructive criticism.